\documentclass[12pt]{article}

\usepackage{amsmath, amssymb}
\usepackage{graphicx}
\usepackage{hyperref}
\usepackage{natbib}
\usepackage{geometry}
\usepackage{booktabs}
\usepackage{multirow}
\usepackage{xcolor}
\usepackage{lineno}
\usepackage{subcaption}
\usepackage{float}
\usepackage{xurl}

\title{Probabilistic Forecasting of Wildfire Size Using Conditional Flow Matching}

\author{
  Isaac Lee${^1}$,
  Bryan Shaddy${^2}$,
  Assad Oberai${^2}$ \\
  \\
  \small ${^1}$ Department of Mathematics, University of Southern California \\
  \small ${^2}$ Department of Aerospace and Mechanical Engineering, University of Southern California
}

\date{}

\begin{document}

\maketitle

\begin{abstract}
In this work we present a probabilistic model for predicting the temporal evolution of wildfire size using conditional generative algorithms, namely flow matching. Rather than predicting spatially resolved fire progression, the proposed model forecasts a time series of fire acreage, substantially reducing computational cost and enabling efficient integration with downstream applications such as early warning systems and risk assessment tools. Given estimates of the fire size at the current and previous forecasting steps, alongside spatially averaged weather conditions, terrain characteristics, and fuel composition, the model predicts the change in fire area over a 3-hour interval. Training data are drawn from 140 coupled atmosphere–wildfire simulations generated using WRF-SFIRE, paired with weather data from the North American Mesoscale (NAM) model. Sensitivity analysis demonstrates that the model captures physically-consistent responses to variations in wind speed, temperature, and relative humidity. Model performance is evaluated against reference solutions from WRF-SFIRE. For single-step predictions, the model achieves a symmetric mean absolute percentage error (sMAPE) of 0.39 relative to the reference WRF-SFIRE solutions. Recursive multi-step predictions over 24 hours are performed for 12 held-out test cases, where the flow matching model achieves a final fire area sMAPE of 0.39 while further providing meaningful quantification of forecast uncertainty.

\end{abstract}

\section{Introduction}

Recent years have seen increasing wildfire activity near urban areas, culminating in significant impacts on health, infrastructure, and economies. Notable occurrences include the 2025 Eaton and Palisades fires in Los Angeles county, which led to the evacuation of 180,000 individuals, destroyed more than 16,000 structures, caused an estimated $250$ billion in damages, and resulted in the death of at least 27 people \citep{qiu2025rising,seydi2025assessment,Danielle_2025}. These calamitous trends are expected to further worsen as climate change provides conditions increasingly conducive to wildfires while the number of homes in the wildland-urban interface (WUI) simultaneously expands \citep{flannigan2013global,burke2021changing}. To better mitigate risks associated with wildfires, there is a need for tools aimed at assessing the potential severity of fires early on in their initial growth. 

Accurate and timely forecasting of wildfire size is important for estimating potential wildfire severity, providing critical information for supporting firefighting operations, planning evacuations, and predicting potential smoke dispersion. Quantifying prediction uncertainty additionally aids in assessment and communication of risks in these settings. While state-of-the-art operational models such as WRF-SFIRE couple atmospheric dynamics with fire spread physics to produce high-fidelity predictions of wildfire growth, their computational cost limits their use for rapid assessment of wildfire growth potential and hinders their use for estimating prediction uncertainty \citep{mandel2011}. These limitations motivate a growing interest in machine learning-based models which can predict fire growth at a fraction of the cost of physics-based models to provide timely estimates of wildfire potential \citep{jain2020, andrianarivony2024}.

Recently, conditional generative models have been shown to be well suited for probabilistic wildfire spread prediction. In \citet{shaddy2026}, a spatially-resolved surrogate model based on conditional flow matching is developed, predicting fire progression on a $128 \times 128$ spatial grid at 25-meter resolution and additionally providing quantification of prediction uncertainty. This model takes as input spatial fields of burned area, wind, relative humidity, temperature, terrain height, and fuel category, and produces an ensemble of spatially resolved fire progression maps over a 3-hour timestep. While useful for predicting the detailed spread of wildfires, this model incurs significant overheads in data processing when assembling the input fields that are necessary to drive the model. This in turn motivates the development of models that are simpler and easier to run.
%the use of more straightforward formulations which aim to estimate fire area as a scalar value.

In \citet{coffield2019}, authors demonstrated the viability of machine learning for wildfire size prediction by training decision tree classifiers to categorize Alaskan boreal fire ignitions into small, medium, and large size classes using vapor pressure deficit and spruce cover fraction. However, classification into coarse size categories precludes use with potential downstream applications where a time-resolved estimate of fire area may be necessary. Another method by \citet{shmuel2023machine} predicts daily growth rates based on meteorological conditions, terrain, and fuel, however this approach provides only deterministic solutions, limiting the ability to quantify prediction uncertainty. In \citep{kc2021surrogate}, a surrogate model for predicting fire area growth over a 5 h  period based on conditions at ignition time is developed. The surrogate model, however, relies on a simplistic mathematical expression which is fitted to simulations from the Spark wildfire spread model, resulting in limited ability to capture the complex non-linear relationships between environmental variables and fire growth. 

\begin{figure}
    \centering
    \includegraphics[width=\linewidth]{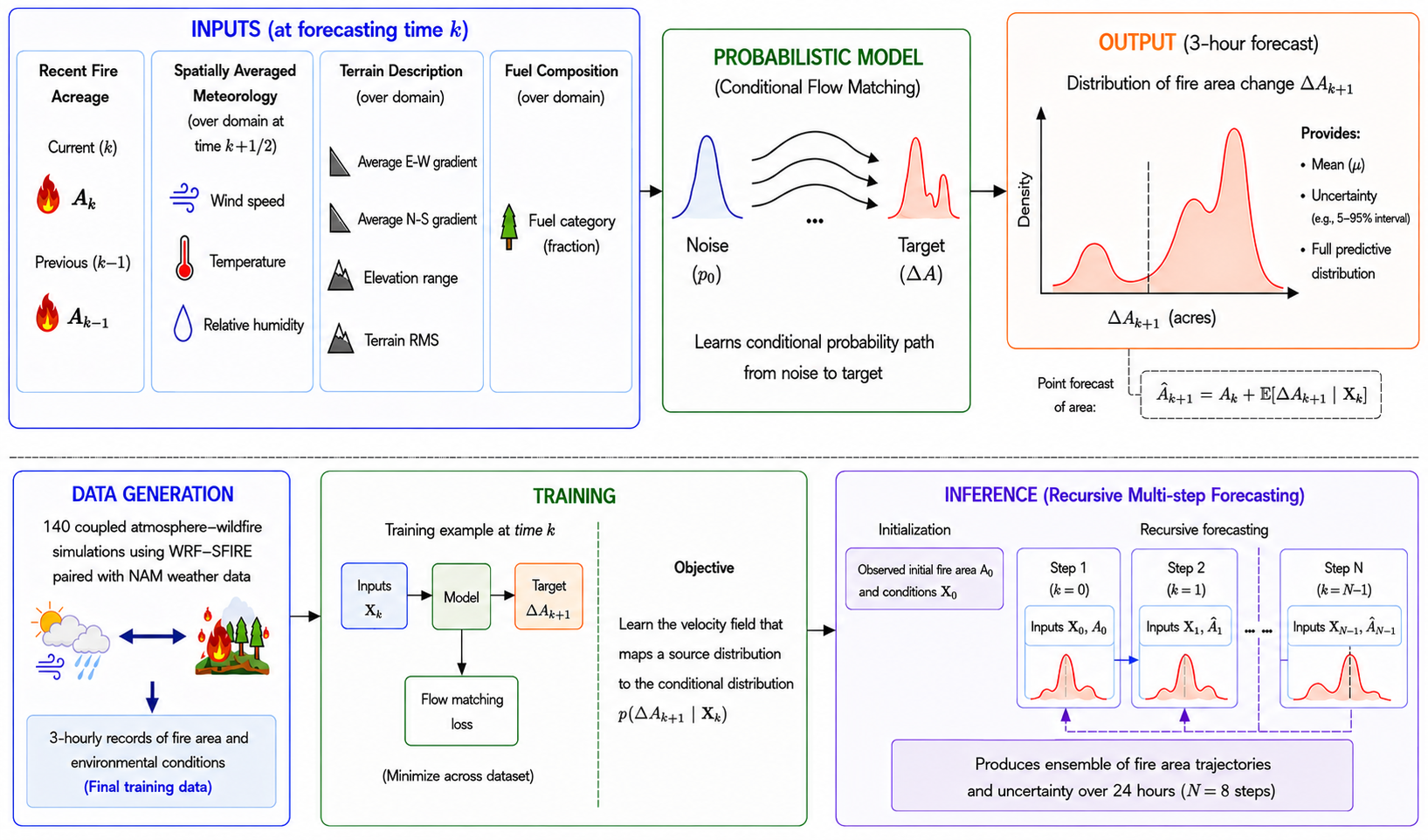}
    \caption{Graphical overview of the flow matching framework for probabilistic forecasting of wildfire size.}
    \label{fig:graphical_overview}
\end{figure}

In this work, we develop a model for probabilistic prediction of the time evolving size of a wildfire. As with previous approaches, rather than predicting a spatially resolved fire arrival time map, our model forecasts the total burned area as a scalar time series based on scalar representations of current fire area, weather, terrain, and fuel variables, using a conditional flow matching framework \citep{lipman2022,dasgupta2026}. The model is trained on wildfire spread solutions generated using WRF-SFIRE, along with corresponding weather data obtained from the North American Mesoscale (NAM) model. Once trained, for a given set of conditional inputs the model generates an ensemble of fire area estimates over a three hour timestep. The model may subsequently be applied in a recursive setting to estimate fire growth over longer time horizons. Figure~\ref{fig:graphical_overview} provides an overview of the modeling framework developed in this work. 

The proposed formulation offers several distinct advantages over other approaches. First, the model is substantially more computationally efficient than spatially resolved physics-based models and machine learning surrogate models, enabling rapid generation of large ensembles at minimal cost thereby enabling field deployment. Second, because the model is trained on data from WRF-SFIRE it is able to represent the complex physical relationships present in this code. Third, by providing probabilistic estimates of fire growth potential, the model is suitable for risk assessment and early warning tasks where quantification of prediction uncertainty is critical. Fourth, by providing a time-series of estimated fire area, the model is directly compatible with downstream applications requiring continuous fire area estimates; for example, plume rise models requiring a time-evolving fire area estimate as input \citep{freitas2007including}. 

After developing and training the model, we evaluate its performance through a sensitivity analysis, assessing the influence of input variables on predicted fire growth. Thereafter we quantify its performance in making single time-step (3-hour) predictions for 1,200 cases selected from the test dataset. Lastly, we apply the model to a recursive 24-hour roll-out of fire growth for 12 held-out wildfire events and compare the results against WRF-SFIRE solutions. Uncertainty estimates from the flow matching-based approach are further assessed by examining correlation with prediction error.

The remainder of this paper is organized as follows. Section~\ref{sec:methods} describes the problem formulation, the conditional flow matching algorithm, and the training data generation procedure. Section~\ref{sec:results} presents results from the sensitivity analysis, single-step validation, and recursive model roll-out. Section~\ref{sec:conclusions} summarizes findings and outlines future directions.

\section{Methods}
\label{sec:methods}
In the following section we provide details on the algorithm developed here for probabilistic prediction of wildfire size based on environmental conditions. We first detail the problem formulation, then provide a description of the underlying flow matching algorithm, after which the approach for the autoregressive roll-out of the algorithm is formalized, and lastly the training data generation procedure is outlined. 

\subsection{Problem Formulation}
\label{sec:formulation}

The goal of this work is to predict the evolution of wildfire size based on weather, fuel, and terrain data. Rather than predicting spatially resolved wildfire progression, we formulate wildfire growth prediction as a scalar regression problem in which both inputs and outputs are low-dimensional vectors of aggregated quantities. 

The fire area prediction model operates on a 3-hour forecasting interval, where at each forecast step the model estimates the change in fire area over that timestep based on both the current burned area and the burned area at the previous timestep. The model additionally takes as input spatially averaged environmental conditions, valid at the interval midpoint. Environmental variables are averaged over a 7.5 km x 7.5 km domain centered on the ignition point for the fire. The model is further formulated to enable initialization either from ignition, or from a known wildfire size, enabling easy application throughout the duration of a fire.

Given the prescribed inputs, the model predicts the change in wildfire area 1, 2, and 3 hours beyond the current timestep. We denote the output vector as $\mathbf{Y}$. The total wildfire area can therefore be obtained by adding the incremental growth predicted by the model to the current fire area passed as input to the model. We note that the fire areas used as input, namely the current fire area and fire area from the previous timestep, refer to the total wildfire area values obtained at the end of each 3 hour timestep interval. Here intermediate predictions after 1 and 2 hours are provided for applications requiring sub-step fire area estimates. 

The input to the model is denoted by the vector $\mathbf{X}$ and consists of the following scalar quantities:
\begin{itemize}
    \item Current burned area and burned area obtained from the previous timestep, providing temporal context for fire growth rate.
    \item Spatially averaged 10-meter zonal and meridional wind components.
    \item Spatially averaged 2-meter relative humidity. 
    \item Spatially averaged 2-meter temperature.
    \item Terrain metrics including the average terrain gradient in the East-West direction, average terrain gradient in the North-South direction, difference between maximum terrain height and minimum terrain height, and terrain RMS roughness.
    \item Fuel composition, determined by the spatial area occupied by each of the 13 Anderson Fire Behavior Fuel Model categories, along with a no fuel category, within the 7.5 km x 7.5 km domain.
\end{itemize}
\noindent The full input vector has dimension $|\mathbf{X}| = 24$ and the output vector has dimension $|\mathbf{Y}|=3$. 

We recognize that due to the inherent chaotic behavior of wildfire spread, along with the incomplete knowledge of the spatial structure of weather, terrain, and fuel conditions used in the problem formulation prescribed here, a probabilistic mapping between inputs $\mathbf{X}$ and outputs $\mathbf{Y}$ is desirable. 

Therefore, we model $\mathbf{X}$ and $\mathbf{Y}$ as random vectors. The objective then becomes to characterize the conditional distribution $P_{\mathbf{Y}|\mathbf{X}}$ such that for a given instance of inputs $\mathbf{X}$ we can draw samples of corresponding wildfire growth estimates $\mathbf{Y}$ from this distribution. 

To obtain samples from the desired conditional density, conditional generative models are utilized. Specifically, we employ conditional flow matching algorithms, following the work of \citep{dasgupta2026,lipman2022,tong2023}. The conditional flow matching model is trained on tuples of $(\mathbf{X},\mathbf{Y})$ from the joint distribution $P_{\mathbf{Y},\mathbf{X}}$, which are derived from simulations from the coupled atmosphere-wildfire model WRF-SFIRE, along with corresponding weather data from the North American Mesoscale Model. 

\subsection{Conditional Flow Matching}
\label{sec:cfm}

To generate samples from the desired conditional distribution $P_{\mathbf{Y} | \mathbf{X}}$, we employ a conditional flow matching (CFM) algorithm following that of \citet{dasgupta2026}. We provide a brief summary of the CFM algorithm here and refer the reader to \citet{lipman2022, tong2023, dasgupta2026} for additional details.

To begin, we define the pseudo-time variable $t \in (0, 1)$ and introduce the random variable $\mathbf{Y}_t$ which is defined for all $t$ as 
\begin{equation}
    \mathbf{Y}_t = \mathbf{Z}(1 - t) + \mathbf{Y} t,
\end{equation}
where $\mathbf{Z} \sim \mathcal{N}(\mathbf{0}, \mathbf{I})$ and $\mathbf{Y} \sim P_{\mathbf{Y} | \mathbf{X}}$. At $t = 0$, $\mathbf{Y}_0 = \mathbf{Z}$ is pure noise, and at $t = 1$, $\mathbf{Y}_1 = \mathbf{Y}$ is a sample from the target distribution. Thus, $\mathbf{Y}_t$ is a mixture of $\mathbf{Z}$ and $\mathbf{Y}$ parameterized by $t$. Corresponding to this definition of $\mathbf{Y}_t$, there exists a velocity field $v_t(\mathbf{Y}_t, \mathbf{X})$ such that if we select samples $\mathbf{Y}_0 \sim P_{\mathbf{Z}}$ and integrate using
\begin{equation}
    \frac{d\mathbf{Y}_t}{dt} = v_t(\mathbf{Y}_t, \mathbf{X})
\end{equation}
from $t = 0$ to $t = 1$, samples $\mathbf{Y}_0 \sim \mathcal{N}(\mathbf{0}, \mathbf{I})$ can be mapped to samples $\mathbf{Y}_1 \sim P_{\mathbf{Y} | \mathbf{X}}$. This velocity field is approximated using a neural network, which is obtained by minimizing the loss
\begin{equation}
    \mathcal{L}(v_t) = \frac{1}{N} \sum_{i=1}^{N} \left| v_{t^{(i)}}\!\left(\mathbf{y}_{t^{(i)}}, \mathbf{x}^{(i)}\right) - \left(\mathbf{y}^{(i)} - \mathbf{z}^{(i)}\right) \right|^2,
\end{equation}
where $t^{(i)} \sim \mathcal{U}(0,1)$, $\mathbf{z}^{(i)} \sim \mathcal{N}(\mathbf{0}, \mathbf{I})$, and $(\mathbf{y}^{(i)}, \mathbf{x}^{(i)}) \sim P_{\mathbf{Y}\mathbf{X}}$. 

\begin{figure}[H]
    \centering
    \includegraphics[width=0.6\textwidth]{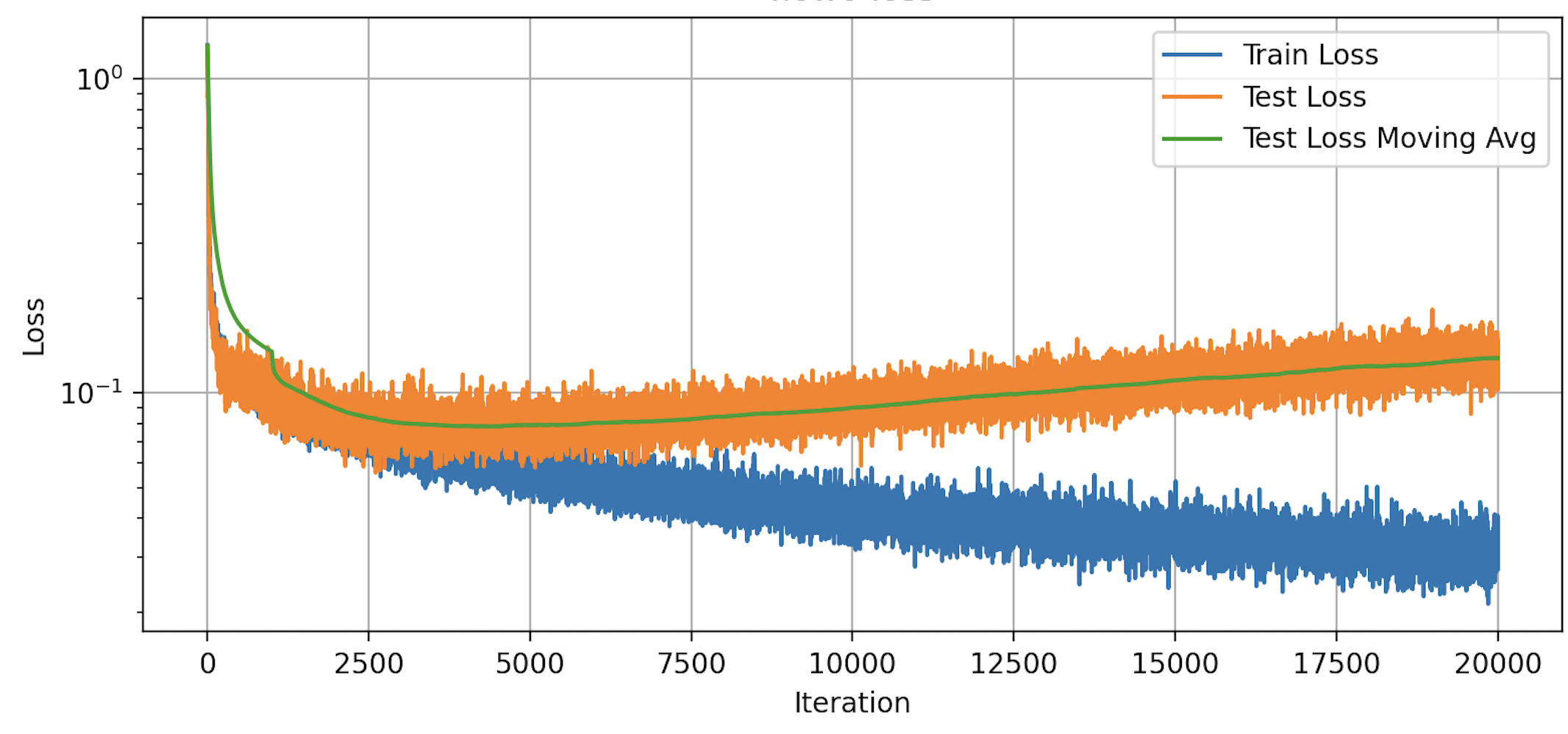}
    \caption{Training and test loss curves for the velocity network of the conditional flow matching model.}
    \label{fig:loss}
\end{figure}

The velocity field is modeled using a fully connected feed-forward neural network consisting of 4 hidden layers with a width of 256. The model is implemented in PyTorch and was trained using a batch size of 2,000 for a total of 20,000 mini-batches, or approximately 2,857 epochs. The network is trained using the Adam optimizer with learning rate $10^{-3}$, and an exponential moving average of model parameters is maintained throughout training with a decay rate of 0.9999. To generate the final results presented, a model checkpoint corresponding to 10,000 training iterations is utilized, which is just past the apparent minimum in the test loss curve, thus ensuring the model was sufficiently trained without risking variance collapse or overfitting. The training and test loss curves are shown in Figure~\ref{fig:loss}.

\subsection{Autoregressive Rollout}
\label{sec:rollout}

For multi-step prediction, the model is applied autoregressively to simulate fire area evolution over time horizons greater than 3 hours. At each forecast step the model receives the current and previous fire area estimates, alongside the corresponding weather, terrain, and fuel inputs for this timestep, and produces an ensemble of predicted fire area growth over a 3-hour interval. The predicted growth increment is then added to the current area used as input to obtain the fire area estimate resulting from the 3-hour growth period, which is subsequently used as input for the following prediction. This procedure is repeated sequentially to produce a trajectory of the desired length.

Uncertainty is propagated through time using an ensemble of $M = 100$ realizations. In the first timestep we begin from a single set of conditioning inputs, which are used to generate our ensemble of predictions. Following this, in subsequent timesteps each ensemble member is used to generate a single prediction for the next timestep, in a one-to-one mapping. This approach ensures the number of ensemble members remains fixed throughout the simulation. Additionally, all other conditioning variables remain fixed across the ensemble of predictions, namely weather, terrain, and fuel variables. This means ensemble members share the same conditioning inputs at each step but represent independent samples from the prior, preserving diversity in the ensemble without requiring explicit perturbation of inputs. The result of each rollout is an ensemble of predicted fire area trajectories of the desired length. 

To obtain a point forecast, the ensemble mean is computed across all members at each 3-hour step. The standard deviation across ensemble members is additionally computed at each 3-hour step and provides a measure of predictive uncertainty.

\subsection{Training Data}
\label{sec:data}

\subsubsection{WRF-SFIRE Simulations}
Training data is derived from wildfire spread simulations generated using the coupled atmosphere-wildfire model WRF-SFIRE. WRF-SFIRE consists of a coupling between the Weather Research and Forecasting model, WRF, and a fire propagation model, SFIRE, which employs a level-set solver with spread rates derived from the Rothermel rate-of-spread model \citep{mandel2011,rothermel1972mathematical}. For running the simulations the wrfxpy system was utilized, which automates the simulation workflow including simulations setup and execution. 

Simulations were generated for 140 wildfires occurring in the contiguous United States (CONUS) during 2023, with 12 additional simulations reserved for testing. Simulations occurred on a domain of size 30 km x 30 km, with a fire grid of 1200 x 1200 utilized to represent fire propagation with a resolution of 25 m. For the atmospheric model, a horizontal resolution of 1 km was utilized, along with 40 vertical levels and a model top of approximately 5000 Pa, corresponding to an altitude of roughly 20 km. 

For initial and boundary conditions for the atmospheric component of WRF, data from the North American Mesoscale (NAM) Forecast System Grid 227 were utilized. The NAM data was available at 5 km resolution, and was collected using a 3-hour interval. Additionally, fuel data was obtained from the 2019 LANDFIRE 13 Anderson Fire Behavior Fuel Models (FMFM13) dataset and data for terrain elevation were derived from the USGS National Elevation Dataset (NED). 

Ignition information was derived from VIIRS and GOES detections provided by NOAA's Next Generation Fire System (NGFS). NGFS clusters active fire detections into incident groups which are associated with fires cataloged by the National Interagency Fire Center (NIFC). Ignition times were estimated based on the timestamp of the earliest detection associated with each fire event. Ignition locations were derived from the average location of all VIIRS detections within the footprint of the initial GOES detection associated with an event; in cases where no VIIRS detections were made within the first four hours following an initial GOES detection, ignition locations were estimated from the average position of available GOES detections. 

Ignitions were placed approximately in the center of the domain, and simulations were run for a duration of 48 hours. Resulting from each simulation were fire arrival time maps, from which fire size can be obtained, along with corresponding NAM weather data, terrain, and fuel information. 

\subsubsection{Training Data Construction}
Construction of training data from the WRF-SFIRE solutions began with a data augmentation procedure consisting of random rotations and random sampling of forecasting times. Data augmentation through rotations was performed to diversify the combinations of directional fields in the training data, while sampling of forecasting times allowed for the inclusion of many starting states in the dataset. During the data augmentation procedure, it is ensured that directional fields such as wind are projected back onto a fixed N-S and E-W coordinate system. For each WRF-SFIRE simulation, we apply 10 random rotations drawn from $\mathcal{U}(0^\circ, 360^\circ)$, and for each rotation we sample 10 forecast times drawn from $\mathcal{U}(T_{\text{ign}}, T_{\text{max}} - \Delta t)$, where $T_{\text{max}}$ is selected to be 24 hours after ignition. The data augmentation procedure yields 100 samples per simulation, and thus 14,000 samples for training along with 1,200 samples for testing.  

From each augmented WRF-SFIRE solution, we extract an input burned area corresponding to the selected forecasting time, along with the corresponding change in area over the forecast step. Weather data for a selected forecasting time is obtained first through linear interpolation in time to obtain values at the midpoint between forecasting steps, $t_n + \Delta t / 2$. Scalar inputs are then derived from the weather data, terrain, and fuel fields through spatial averaging and aggregation, following what is outlined in Section~\ref{sec:formulation}. Spatial averages are computed over the central $300 \times 300$ pixels (7.5 km $\times$ 7.5 km) of the WRF-SFIRE domain, which is guaranteed to contain valid non-zero values under all data augmentation rotations. Terrain gradients are computed using finite differences at 25-meter grid spacing and averaged over the same region. 

With the desired scalar input values obtained, they are then normalized to be in the range $[0,1]$ as follows. Wind components are normalized by adding 13.5 and dividing by 27. Relative humidity is normalized by dividing by 100. Temperature is normalized by subtracting 260 and dividing by 60. Terrain gradients are normalized by adding 0.14 and dividing by 0.28, maximum terrain variation is divided by 2500 m, and terrain RMS roughness is divided by 425 m. Fuel category pixel counts are normalized by the total number of pixels in the $300 \times 300$ region (90,000 pixels). Current and previous burned areas are normalized by 19,000 and 14,000 acres, respectively, reflecting the approximate range of values observed in the training data.

To normalize the output values for training, fire growth area is normalized using a logarithmic transformation. The distribution of fire growth increments is heavily right-skewed: the distribution of 3-hour fire area growth in the training set has a mean of 384 acres, a median of 148 acres, a 90th percentile of 954 acres, and a maximum of 13,100 acres. This wide range and heavy tail mean that standard linear normalization concentrates most of the probability mass near zero and can lead to physically implausible negative predicted increments during inference. We address this by applying the transformation
\begin{equation}
    \hat{Y}_i = \frac{\ln(Y_i + 1)}{\ln(Y_i^{\max} + 1)},
\end{equation}
where $Y_i, \; i = 1,2,3$ is a single component of the fire area growth vector $\mathbf{Y}$, $Y_i^{\max}$ is the maximum value of $Y_i$ across the training set, computed separately for each output index $i = 1, 2, 3$, and $\hat{Y}_i$ is the normalized fire growth area value for component $i$. This transformation maps all non-negative increments to the interval $[0, 1]$, preserves the ordering of growth magnitudes, and compresses the heavy tail of the distribution. Crucially, since $\ln(x + 1) \geq 0$ for all $x \geq 0$, the transformation enforces non-negativity in the normalized space. During inference, predicted normalized increments are mapped back to physical units by inverting the transformation. 

\section{Results}
\label{sec:results}

\subsection{Sensitivity Analysis}
To assess the performance of the wildfire growth prediction model, we first evaluate whether physically meaningful relationships between input variables and predicted fire growth are being captured. We begin by prescribing a controlled set of baseline conditions, following which each input is independently varied to evaluate the sensitivity of the model to that variable. For each set of conditioning inputs we generate 1000 samples, from which the mean and standard deviation of predicted fire area growth over the 3-hour interval are estimated.

The baseline configuration consisted of zero wind ($\bar{u} = \bar{v} = 0$ m s$^{-1}$), a relative humidity of 20\%, a temperature of approximately 300 K ($\approx$80$^\circ$F), a flat terrain (all terrain metrics set to zero), and a single fuel type corresponding to Anderson fuel category 9 (hardwood litter). Current and previous fire areas are both set to zero to remove the influence of growth trends and isolate the effects of environmental variables. We vary the zonal wind component $\bar{u}$ over $[-5, 5]$ m s$^{-1}$ and the meridional wind component $\bar{v}$ over the same range, each independently while holding all other inputs fixed at the baseline. Relative humidity is varied from 0\% to 70\%, and temperature is varied from 46.5$^\circ$F to 87$^\circ$F in increments of 5 K. Roughly 80\% of the training data falls within $\pm 5$ m s$^{-1}$ for each wind component, between 0\% and 70\% for relative humidity, and between 46.5$^\circ$F and 87$^\circ$F ($\approx$281--304 K) for temperature, indicating that the sensitivity curves are evaluated primarily over the well-represented region of the training distribution. Terrain variables are excluded from this analysis because the scalar terrain metrics (average gradients, maximum height variation, and RMS roughness) encode combined geometric effects that are difficult to vary independently in a physically interpretable way.

\begin{figure}[H]
    \centering
    \begin{subfigure}[b]{0.44\textwidth}
        \centering
        \includegraphics[width=\textwidth]{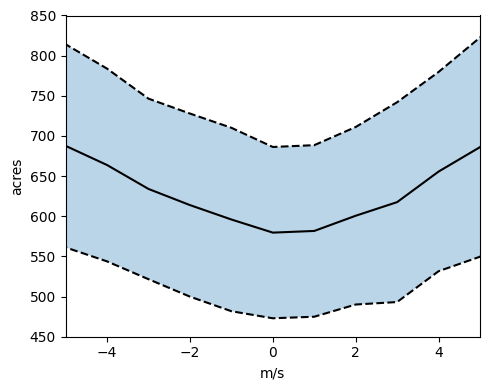}
        \caption{U-Wind}
    \end{subfigure}
    \hfill
    \begin{subfigure}[b]{0.44\textwidth}
        \centering
        \includegraphics[width=\textwidth]{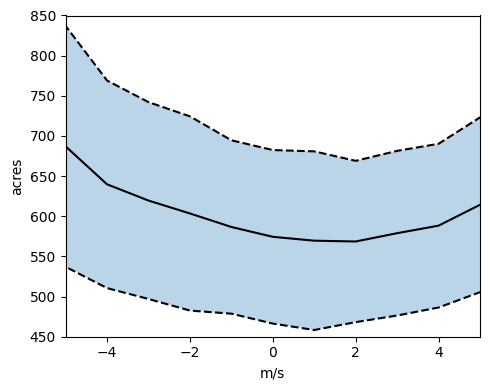}
        \caption{V-Wind}
    \end{subfigure}

    \vspace{0.2em}

    \begin{subfigure}[b]{0.44\textwidth}
        \centering
        \includegraphics[width=\textwidth]{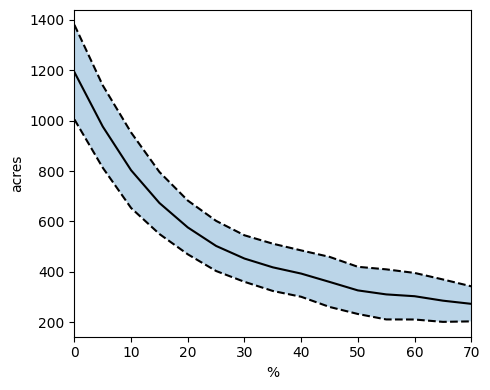}
        \caption{Relative Humidity}
    \end{subfigure}
    \hfill
    \begin{subfigure}[b]{0.44\textwidth}
        \centering
        \includegraphics[width=\textwidth]{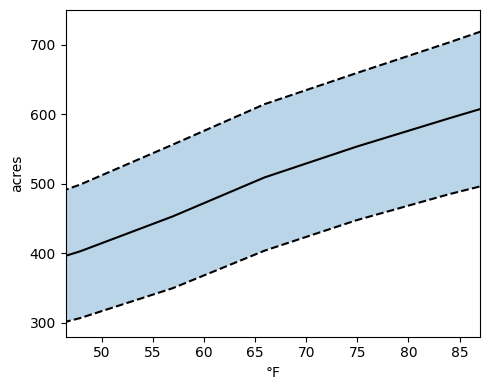}
        \caption{Temperature}
    \end{subfigure}

    \caption{Predicted fire growth area over a single timestep for varying (a) u-wind, (b) v-wind, (c) relative humidity, and (d) temperature. For each plot, the mean predicted growth area (solid line) and mean $\pm 1$ standard deviation (shaded band) are indicated.}
    \label{fig:sensitivity}
\end{figure}

Results for the sensitivity analysis are presented in Figure~\ref{fig:sensitivity}, from which we observe that the model produces physically reasonable responses across all environmental variables examined. Predicted fire growth increases with wind magnitude for both wind components, with a minimum near zero wind speed. We do note that there is a slight asymmetry in the observed response curves, though the general desired trends are captured. Predicted area additionally decreases monotonically with increasing relative humidity, reflecting the known dampening effect of atmospheric moisture on fire spread. Predicted fire growth is also observed to increase with temperature, consistent with expected behavior. These relationships provide evidence that the scalar surrogate captures the dominant physical drivers of wildfire growth, even without information of spatial structure of environmental conditions. %We note that the absolute magnitude of predicted fire area growth in the sensitivity analysis is somewhat elevated relative to the training data mean, which we attribute to the sensitivity baseline prescribing zero current and previous fire area, which is a configuration that is underrepresented in the training distribution and for which the model extrapolates toward larger values. The sensitivity analysis is therefore interpreted primarily in terms of the qualitative relationships between input variables and predicted growth, rather than the absolute magnitudes.

\subsection{Single-Step Validation}

To further assess the model, we evaluate single-step (3-hour) prediction performance on the 1,200 held-out test cases. For each case, we generate 1000 stochastic samples using the flow matching model and compute the ensemble mean as a point estimate. We report prediction accuracy using the symmetric mean absolute percentage error (sMAPE), defined as
\begin{equation}
    \text{sMAPE} = \frac{1}{N} \sum_{i=1}^{N} \frac{|y_i - \hat{y}_i|}{(|y_i| + |\hat{y}_i|)/2},
\end{equation}
where $y_i$ is the target fire area at the end of the 3-hour interval and $\hat{y}_i$ is the predicted value. Here we choose to utilize the sMAPE as our error metric because it is well defined when target values are near zero. We additionally assess the coefficient of determination ($R^2$) to further evaluate the agreement between predicted and target fire growth area values. 

\begin{figure}[H]
    \centering
    \begin{subfigure}[b]{0.48\textwidth}
        \centering
        \includegraphics[width=\textwidth]{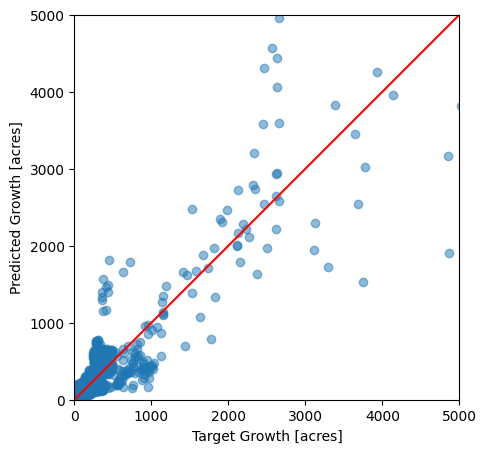}
        \caption{Full range}
    \end{subfigure}
    \hfill
    \begin{subfigure}[b]{0.48\textwidth}
        \centering
        \includegraphics[width=\textwidth]{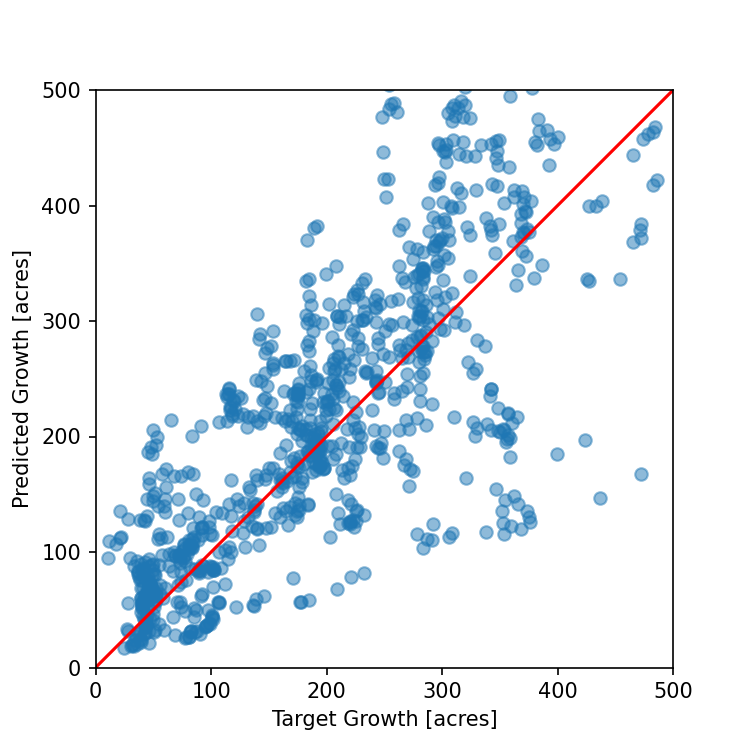}
        \caption{Zoomed view}
    \end{subfigure}
    \caption{Predicted versus target burned area for a single-step (3-hour) prediction using the conditional flow matching model. Blue points indicate individual cases from the test set and the red line indicates perfect agreement. The right panel shows a zoomed view of the region of highest data density.}
    \label{fig:singlestep}
\end{figure}

Evaluating the performance of the conditional flow matching model on the test set, we find that a sMAPE of 0.39 and $R^2$ of 0.56 are achieved. A scatter plot of predicted versus target fire area growth for the flow matching model is shown in Figure~\ref{fig:singlestep}.

\begin{figure}[H]
    \centering
    \begin{subfigure}[b]{0.48\textwidth}
        \centering
        \includegraphics[width=\textwidth]{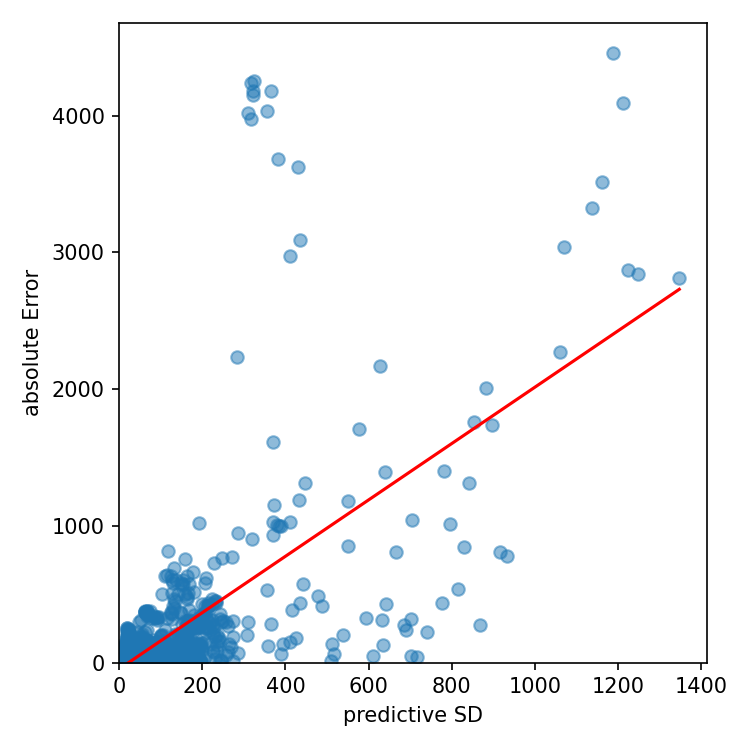}
        \caption{Full range}
    \end{subfigure}
    \hfill
    \begin{subfigure}[b]{0.48\textwidth}
        \centering
        \includegraphics[width=\textwidth]{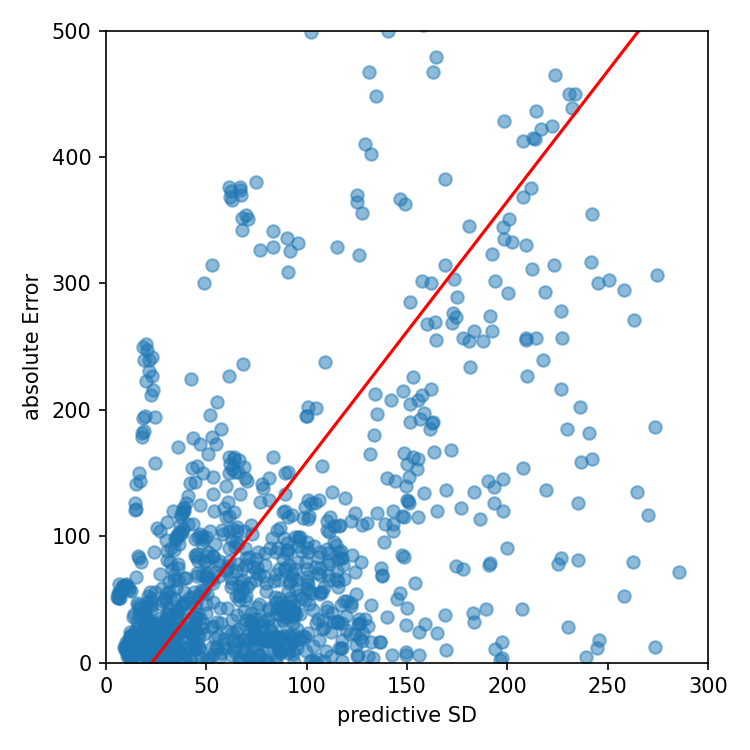}
        \caption{Zoomed view}
    \end{subfigure}
    \caption{Ensemble standard deviation versus absolute prediction error for the 1,200 single-step test cases. The Pearson correlation coefficient is $r = 0.64$ ($p < 0.001$).}
    \label{fig:pearson}
\end{figure}

To assess the quality of the uncertainty estimates provided by the flow matching model, we examine the ensemble standard deviation against the prediction error for each of the 1,200 test cases. Figure~\ref{fig:pearson} depicts absolute prediction error as a function of ensemble standard deviation for each of the 1,200 test cases; a linear regression fit is additionally provided to highlight the relationship between these variables. From Figure~\ref{fig:pearson} we see that in general there is a meaningful relationship between prediction uncertainty and error, with higher uncertainty tending to correspond to larger errors. We further compute the Pearson correlation coefficient between the ensemble standard deviation and the absolute prediction error across the 1,200 test cases, finding a value of $r = 0.64$ ($p < 0.001$), again indicating that when error is large the model tends to indicate higher uncertainty. Additionally, in 54\% of the 1,200 test cases examined, the target value lies within one standard deviation of the mean.

\begin{figure}[H]
    \centering
    % Row 1
    \begin{subfigure}[b]{0.48\textwidth}
        \includegraphics[width=\textwidth]{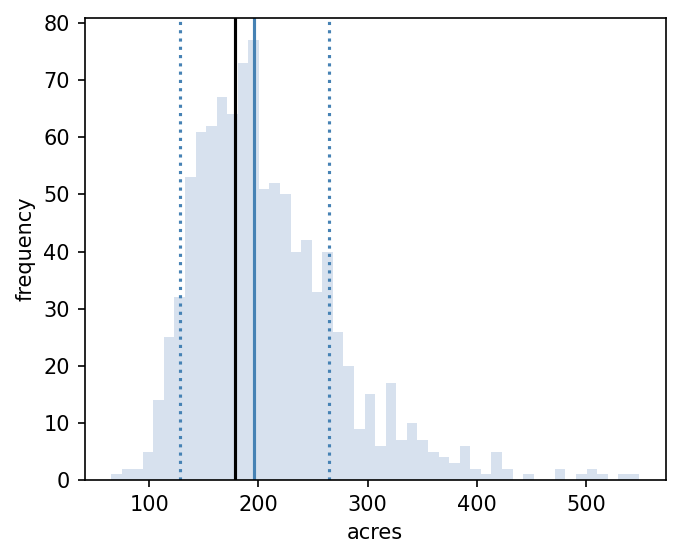}
        \caption{Sample 1}
    \end{subfigure}
    \hfill
    \begin{subfigure}[b]{0.48\textwidth}
        \includegraphics[width=\textwidth]{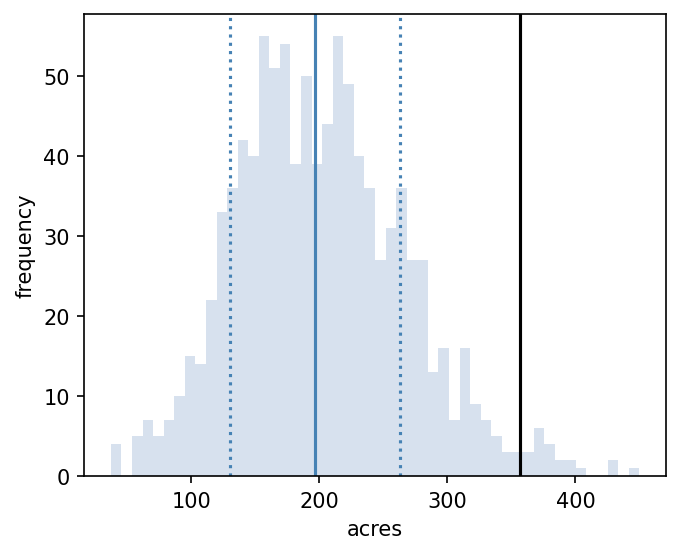}
        \caption{Sample 2}
    \end{subfigure}
    % Row 2
    \begin{subfigure}[b]{0.48\textwidth}
        \includegraphics[width=\textwidth]{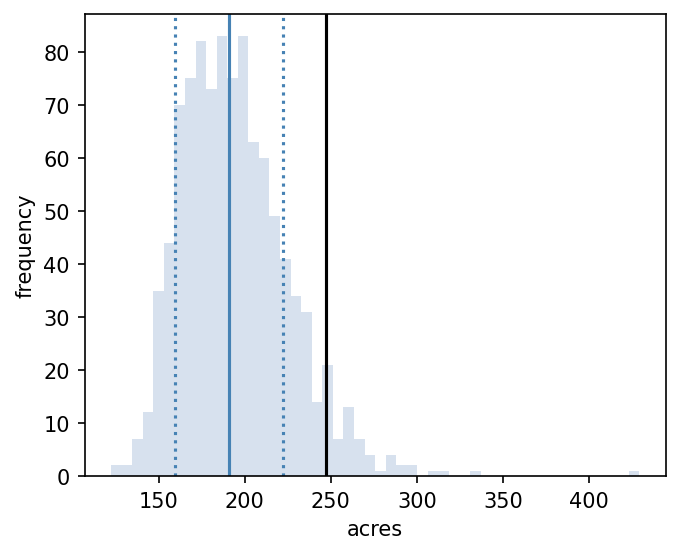}
        \caption{Sample 3}
    \end{subfigure}
    \hfill
    \begin{subfigure}[b]{0.48\textwidth}
        \includegraphics[width=\textwidth]{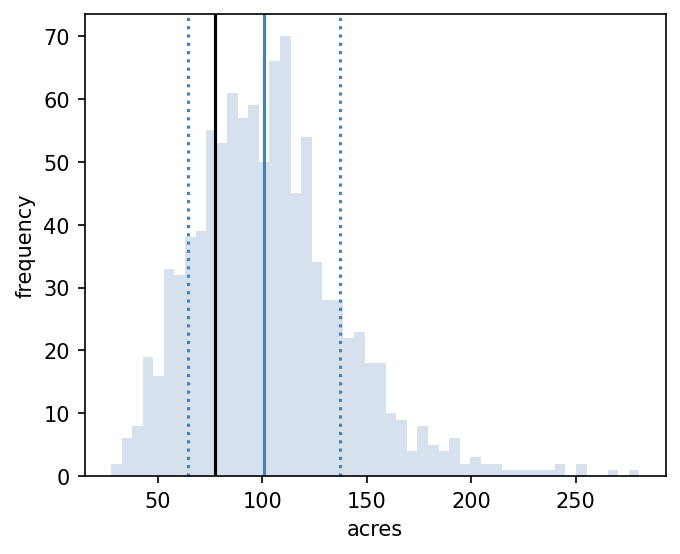}
        \caption{Sample 4}
    \end{subfigure}
    % Row 3
    \begin{subfigure}[b]{0.48\textwidth}
        \includegraphics[width=\textwidth]{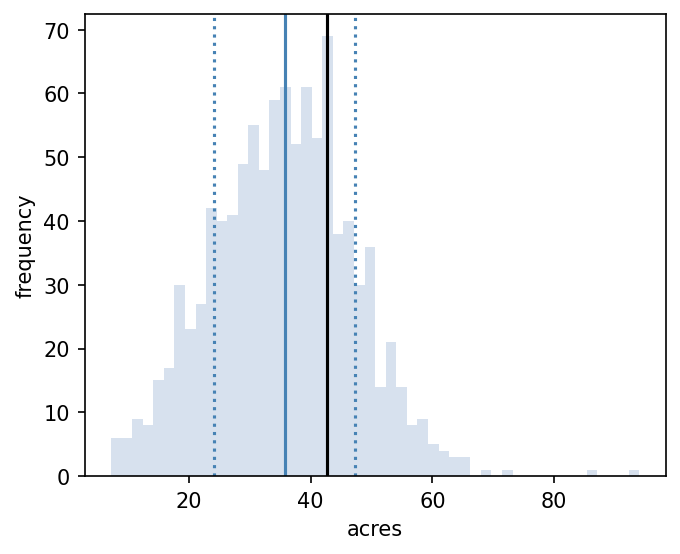}
        \caption{Sample 5}
    \end{subfigure}
    \hfill
    \begin{subfigure}[b]{0.48\textwidth}
        \includegraphics[width=\textwidth]{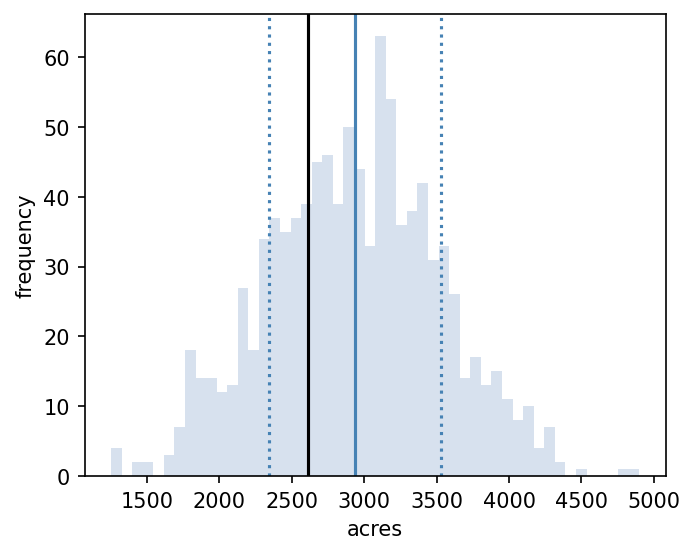}
        \caption{Sample 6}
    \end{subfigure}
    % Legend
    \vspace{0.5em}
    \begin{subfigure}[b]{\textwidth}
        \centering
        \includegraphics[width=0.8\textwidth]{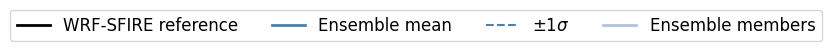}
    \end{subfigure}
    \vspace{-2.5em}
    \caption{Single-step (3-hour) predicted fire area growth distributions for six randomly selected test cases, with the mean predicted value, mean $\pm$ standard deviation range, and target values from WRF-SFIRE indicated.}
    \label{fig:singlestep_hist}
\end{figure}

To further illustrate the probabilistic nature of the flow matching model, Figure~\ref{fig:singlestep_hist} shows predicted fire growth area distributions for six randomly selected test cases. In each case, the ensemble of 1000 samples provides a distribution of possible fire area growth, with the ensemble mean serving as the point estimate. Distribution spread is additionally assessed using the mean $\pm$ standard deviation range. Sample 6 exhibits particularly large fire growth, which the model appears to capture well; inspection of the conditioning inputs for this case reveals relatively high wind speed, low relative humidity, and high temperature, which are conditions strongly associated with rapid fire spread. In samples 2 and 3, the model underpredicts fire growth and the reference value falls outside the ensemble spread. These cases are characterized by relatively low wind speeds and low temperatures.

\subsection{Recursive 24-Hour Rollouts}
\label{sec:rollouts}

We evaluate performance of the wildfire size prediction model for extended time horizons through an autoregressive application of the model over 8 timesteps for the 12 held-out test cases. Growth is predicted starting from ignition out to 24 hours, following the procedure outlined in Section~\ref{sec:rollout}. Predicted fire area growth trajectories are then evaluated against target solutions generated by WRF-SFIRE. 

\begin{figure}[H]
    \centering
    % Row 1
    \begin{subfigure}[b]{0.48\textwidth}
        \includegraphics[width=\textwidth]{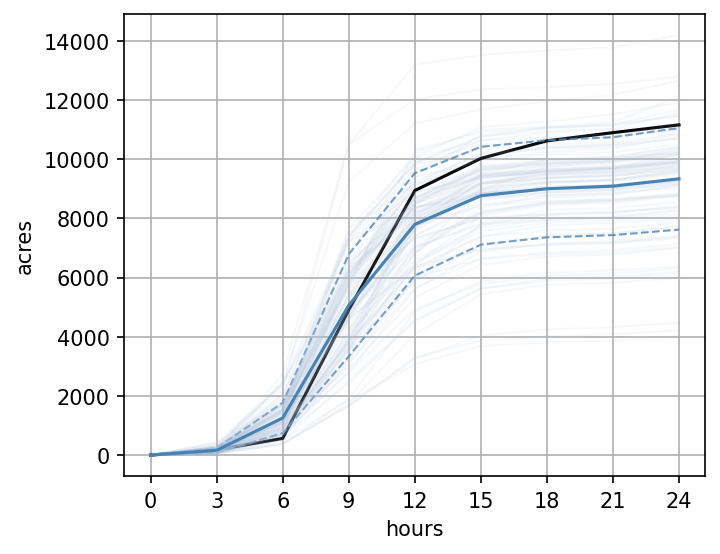}
        \caption{West Hemp}
        \label{fig:rollout_west_hemp}
    \end{subfigure}
    \hfill
    \begin{subfigure}[b]{0.48\textwidth}
        \includegraphics[width=\textwidth]{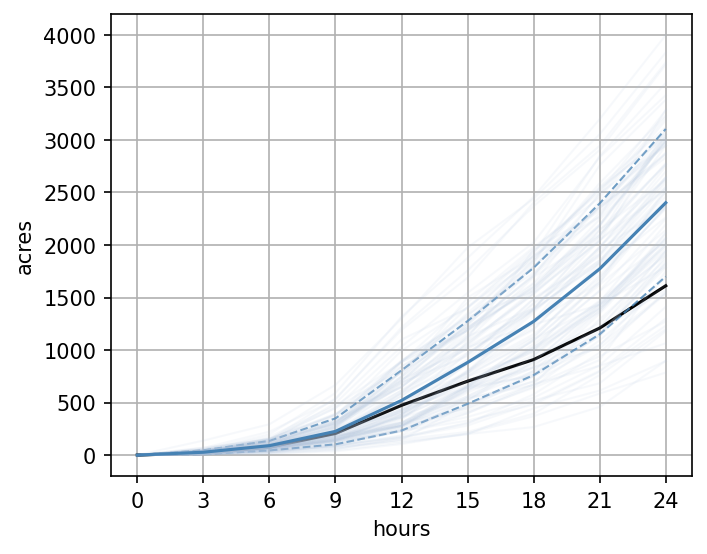}
        \caption{Whiskey}
        \label{fig:rollout_whiskey}
    \end{subfigure}
    % Row 2
    \begin{subfigure}[b]{0.48\textwidth}
        \includegraphics[width=\textwidth]{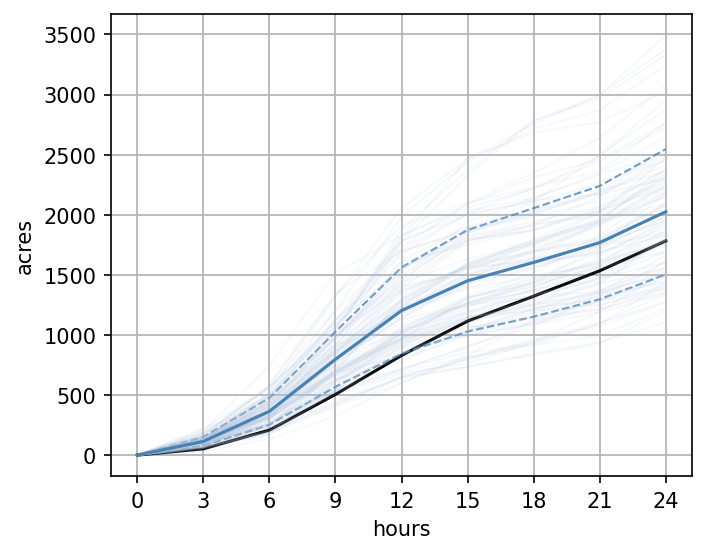}
        \caption{Tyson}
        \label{fig:rollout_tyson}
    \end{subfigure}
    \hfill
    \begin{subfigure}[b]{0.48\textwidth}
        \includegraphics[width=\textwidth]{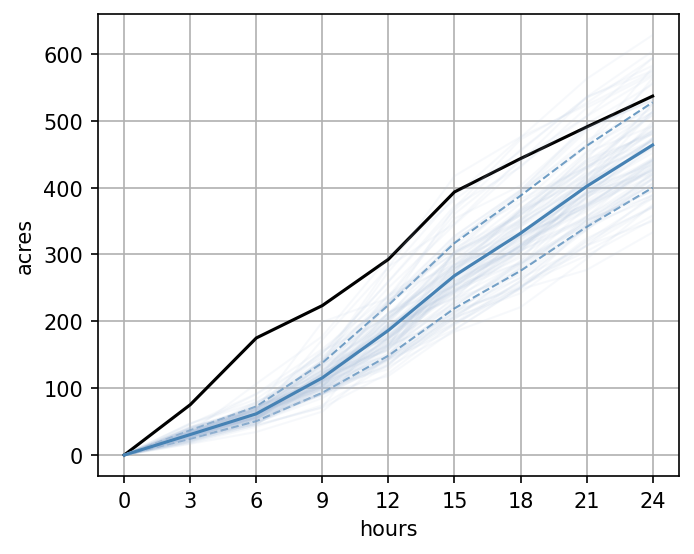}
        \caption{West Fork}
        \label{fig:rollout_west_fork}
    \end{subfigure}
    % Row 3
    \begin{subfigure}[b]{0.48\textwidth}
        \includegraphics[width=\textwidth]{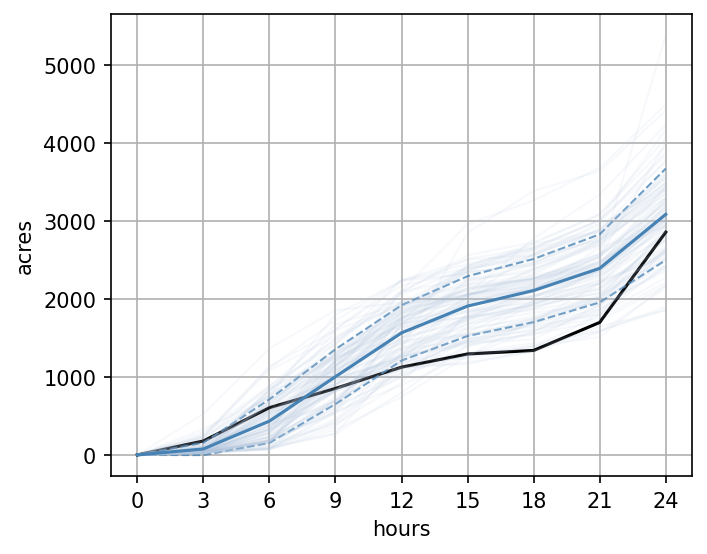}
        \caption{Wildcat}
        \label{fig:rollout_wildcat}
    \end{subfigure}
    \hfill
    \begin{subfigure}[b]{0.48\textwidth}
        \includegraphics[width=\textwidth]{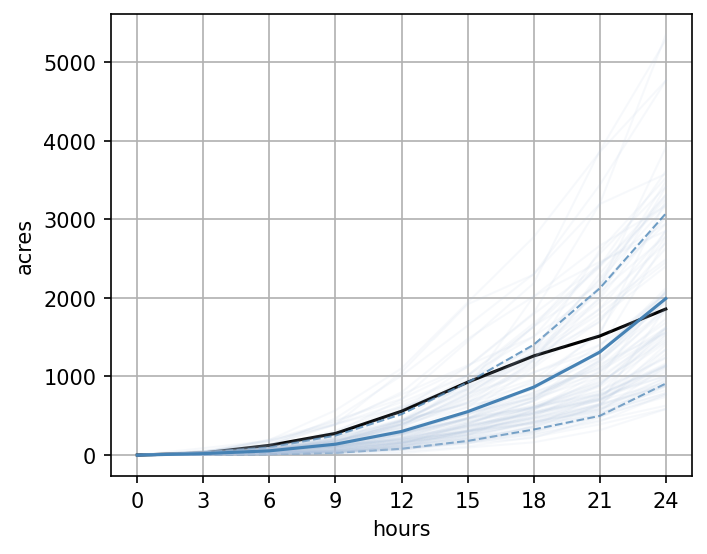}
        \caption{Winters Lake}
        \label{fig:rollout_winters_lake}
    \end{subfigure}
    % Legend
    \vspace{0.5em}
    \begin{subfigure}[b]{\textwidth}
        \centering
        \includegraphics[width=0.8\textwidth]{figures/recursive_flow/legend.png}
    \end{subfigure}
    \vspace{-2em}
    \caption{24-hour fire size predictions for six test fires obtained through recursive application of the conditional flow matching model. Indicated are the 100 ensemble members generated for each case, along with the mean trajectories, mean $\pm$ standard deviation curves, and the reference WRF-SFIRE solutions.}
    \label{fig:rollout_all}
\end{figure}

For each of the 12 test cases, the flow matching model is used to generate an ensemble of 100 24-hour trajectories. From the ensemble of fire size trajectories, the ensemble mean and standard deviation are computed at each 3-hour step, providing a best guess for the size progression for each fire, along with a measure of the ensemble spread. Results for six representative cases are shown in Figures~\ref{fig:rollout_west_hemp}--\ref{fig:rollout_winters_lake}.

From Figure~\ref{fig:rollout_all}, we observe that in most cases the ensemble mean trajectory closely tracks the WRF-SFIRE reference solution over the full 24-hour horizon, often lying in the mean $\pm$ standard deviation range. In cases such as West Hemp, we even see that the model accurately captures sudden, explosive growth periods where the fire expands by thousands of acres over multiple consecutive timesteps. There are instances however where the target solution is outside of the mean $\pm$ standard deviation range, such as West Fork, though we note that in this case the fire is quite small and the resulting solution at the end of the 24 hour prediction period is off by less than 100 acres. 

Examining the mean predicted fire size at the end of the 24 hour growth periods across the 12 test cases, we find an average sMAPE relative to the WRF-SFIRE solutions of 0.39, similar to what was observed for the single-step predictions. This demonstrates the robust ability of the model to predict fire growth even over extended time horizons consisting of multiple applications of the model.

\begin{table}[h]
    \centering
    \caption{Summary of model performance for single-step predictions and recursive rollouts.}
    \label{tab:rollout_metrics}
    \begin{tabular}{lc}
        \toprule
        Metric & Flow Matching \\
        \midrule
        sMAPE (single step) & 0.39 \\
        $R^2$ (single step) & 0.56 \\
        Pearson's Coefficient (single step) & 0.64 \\
        Coverage (mean $\pm$ 1 SD, single step) & 54\% \\
        sMAPE (24-hour final fire area) & 0.39 \\
        \bottomrule
    \end{tabular}
\end{table}

\section{Conclusions and Future Work}
\label{sec:conclusions}

In this work, we have developed a probabilistic model for generating timeseries of wildfire size estimates based on scalar representations of environmental conditions. The model is based on conditional flow matching algorithms, with training data derived from wildfire spread solutions from WRF-SFIRE, along with corresponding weather data from NAM. The model takes as input the burned area at the current timestep and the burned area from the previous timestep, alongside spatially aggregated weather, terrain, and fuel information, and predicts fire area growth over 3-hour intervals. Applied autoregressively, the model can be used to produce 24-hour trajectories of fire area with associated ensemble uncertainty estimates.

% A key methodological contribution is the use of logarithmic normalization for output fire area increments, which enforces physical non-negativity, accounts for the heavy-tailed distribution of fire growth, and leads to measurable improvements in predictive accuracy. This normalization strategy may be broadly applicable to other geophysical prediction problems with similar distributional characteristics.

For single-step predictions, the model achieves a symmetric mean absolute percentage error (sMAPE) of 0.39 relative to the reference WRF-SFIRE solutions. Recursive multi-step predictions over 24 hours are performed for 12 held-out test cases, where the flow matching model achieves a final fire area sMAPE of 0.39, while further providing meaningful quantification of forecast uncertainty.

The scalar formulation offers practical advantages over spatially resolved wildfire spread models. In particular the computational cost is substantially lower, enabling rapid generation of large ensembles with negligible overhead, making this tool suitable for quick assessment of potential wildfire severity. The model is also directly compatible with downstream applications that consume scalar fire area estimates, for example plume rise models. 

Several directions for future work are of interest. Of primary interest is integration of this tool into early warning systems or wildfire risk assessment tools, where probabilistic estimates of potential fire size could be utilized to improve current systems. Additionally, integration with plume rise models is of interest as systems such as this are capable of enabling computationally-efficient generation of ensembles of plume rise estimates, which are broadly of interest. 

\section*{Acknowledgments}

[Acknowledgments to be added.]

\bibliographystyle{plainnat}
\bibliography{references}

@article{mandel2011,
  author    = {Mandel, J. and Beezley, J. and Kochanski, A.},
  title     = {Coupled atmosphere-wildland fire modeling with {WRF-FIRE} version 3.3},
  journal   = {Geoscientific Model Development Discussions},
  volume    = {4},
  number    = {1},
  pages     = {497--545},
  year      = {2011}
}

@article{jain2020,
  author    = {Jain, P. and Coogan, S. C. and Subramanian, S. G. and Crowley, M. and Taylor, S. and Flannigan, M. D.},
  title     = {A review of machine learning applications in wildfire science and management},
  journal   = {Environmental Reviews},
  volume    = {28},
  number    = {4},
  pages     = {478--505},
  year      = {2020}
}

@article{andrianarivony2024,
  author    = {Andrianarivony, H. S. and Akhloufi, M. A.},
  title     = {Machine learning and deep learning for wildfire spread prediction: A review},
  journal   = {Fire},
  volume    = {7},
  number    = {12},
  pages     = {482},
  year      = {2024}
}

@article{lipman2022,
  author    = {Lipman, Y. and Chen, R. T. and Ben-Hamu, H. and Nickel, M. and Le, M.},
  title     = {Flow matching for generative modeling},
  journal   = {arXiv preprint arXiv:2210.02747},
  year      = {2022}
}

@article{tong2023,
  author    = {Tong, A. and Malkin, N. and Huguet, G. and Zhang, Y. and Rector-Brooks, J. and Fatras, K. and Wolf, G. and Bengio, Y.},
  title     = {Conditional flow matching: Simulation-free dynamic optimal transport},
  journal   = {arXiv preprint arXiv:2302.00482},
  volume    = {2},
  number    = {3},
  year      = {2023}
}

@article{dasgupta2026,
  author    = {Dasgupta, A. and Fardisi, A. and Aminy, M. and Binder, B. and Shaddy, B. and Oberai, A.},
  title     = {Solving physics-constrained inverse problems with conditional flow matching},
  journal   = {arXiv preprint arXiv:2603.14135},
  year      = {2026}
}

@article{shaddy2026,
  author    = {Shaddy, B. and others},
  title     = {Generative algorithms for wildfire progression reconstruction from multi-modal satellite active fire measurements and terrain height},
  journal   = {Remote Sensing},
  volume    = {18},
  number    = {2},
  pages     = {227},
  year      = {2026}
}

@article{freitas2007including,
  title={Including the sub-grid scale plume rise of vegetation fires in low resolution atmospheric transport models},
  author={Freitas, Saulo R and Longo, Karla M and Chatfield, Robert and Latham, D and Silva Dias, Maria Assun{\c{c}}{\~a}o Faus da and Andreae, MO and Prins, E and Santos, JC and Gielow, R and Carvalho Jr, JA},
  journal={Atmospheric Chemistry and Physics},
  volume={7},
  number={13},
  pages={3385--3398},
  year={2007},
  publisher={Copernicus Publications G{\"o}ttingen, Germany}
}

@article{coffield2019,
  author    = {Coffield, Shane R. and Graff, Casey A. and Chen, Yang and Smyth, Padhraic and Foufoula-Georgiou, Efi and Randerson, James T.},
  title     = {Machine learning to predict final fire size at the time of ignition},
  journal   = {International Journal of Wildland Fire},
  year      = {2019},
  volume    = {28},
  number    = {11},
  pages     = {861--873},
  doi       = {10.1071/WF19023}
}

@article{seydi2025assessment,
  title={Assessment of the January 2025 Los Angeles County wildfires: A multi-modal analysis of impact, response, and population exposure},
  author={Seydi, Seyd Teymoor},
  journal={arXiv preprint arXiv:2501.17880},
  year={2025}
}

@misc{Danielle_2025, 
  title={AccuWeather estimates more than \$250 billion in damages and economic loss from {LA} wildfires}, 
  author={Danielle, Monica}, 
  howpublished={\url{https://www.accuweather.com/en/weather-news/accuweather-estimates-more-than-250-billion-in-damages-and-economic-loss-from-la-wildfires/1733821}}, 
  publisher={AccuWeather}, 
  year={2025},
  month={Jan}
}

@article{qiu2025rising,
  title={The rising threats of wildland-urban interface fires in the era of climate change: The Los Angeles 2025 fires},
  author={Qiu, Minghao and Chen, Deyang and Kelp, Makoto and Li, Jing and Huang, Guanyu and Yazdi, Mahdieh Danesh},
  journal={The Innovation},
  volume={6},
  number={5},
  year={2025},
  publisher={Elsevier}
}

@article{burke2021changing,
  title={The changing risk and burden of wildfire in the United States},
  author={Burke, Marshall and Driscoll, Anne and Heft-Neal, Sam and Xue, Jiani and Burney, Jennifer and Wara, Michael},
  journal={Proceedings of the National Academy of Sciences},
  volume={118},
  number={2},
  pages={e2011048118},
  year={2021},
  publisher={National Acad Sciences}
}

@article{flannigan2013global,
  title={Global wildland fire season severity in the 21st century},
  author={Flannigan, Mike and Cantin, Alan S and De Groot, William J and Wotton, Mike and Newbery, Alison and Gowman, Lynn M},
  journal={Forest Ecology and Management},
  volume={294},
  pages={54--61},
  year={2013},
  publisher={Elsevier}
}

@article{shmuel2023machine,
  title={A machine-learning approach to predicting daily wildfire expansion rate},
  author={Shmuel, Assaf and Heifetz, Eyal},
  journal={Fire},
  volume={6},
  number={8},
  pages={319},
  year={2023},
  publisher={MDPI}
}

@article{kc2021surrogate,
  title={A surrogate model for rapidly assessing the size of a wildfire over time},
  author={Kc, Ujjwal and Aryal, Jagannath and Hilton, James and Garg, Saurabh},
  journal={Fire},
  volume={4},
  number={2},
  pages={20},
  year={2021},
  publisher={MDPI}
}

@book{rothermel1972mathematical,
  title={A mathematical model for predicting fire spread in wildland fuels},
  author={Rothermel, Richard C},
  year={1972},
  publisher={Research Paper, INT-115. US Department of Agriculture, Intermountain Forest and Range Experiment Station}
}

\subsection*{Data availability}
The data and code supporting the findings of this study are publicly available at \url{https://github.com/isaaclee2/IJWF_Wildfire_Data_and_Model}.

\end{document}